# Deep Generative Models for Relational Data with Side Information

**Changwei Hu** [1]  **Piyush Rai** [2]  **Lawrence Carin** [3]

## Abstract

We present a probabilistic framework for overlapping community discovery and link prediction for relational data, given as a graph. The proposed framework has: (1) a deep architecture which enables us to infer *multiple layers* of latent features/communities for each node, providing superior link prediction performance on more complex networks and better interpretability of the latent features; and (2) a regression model which allows directly conditioning the node latent features on the side information available in form of *node attributes*. Our framework handles both (1) and (2) via a clean, unified model, which enjoys full local conjugacy via data augmentation, and facilitates efficient inference via closed form Gibbs sampling. Moreover, inference cost scales in the number of edges which is attractive for massive but sparse networks. Our framework is also easily extendable to model weighted networks with count-valued edges. We compare with various state-of-the-art methods and report results, both quantitative and qualitative, on several benchmark data sets.

## 1. Introduction

Statistical modeling of complex real-world networks is an important problem, drawing attention from diverse domains, such as social network analysis, biology, political science, etc. (Fortunato, 2010; Goldenberg et al.; Schmidt & Morup, 2013). The goal in statistical modeling of networks is usually to discover the underlying groups or community structure in the network, and/or predicting the existence of potential links between nodes. A common way of accomplishing this is by embedding the nodes in a latent space via latent space models (Hoff et al., 2002). Extensions of the latent space model include stochastic blockmodels (Nowicki & Snijders, 2001), and variants thereof (Miller et al., 2009; Airoldi et al., 2008; Latouche et al., 2011), which can learn node embeddings that are interpretable (e.g., sparse) and can therefore reflect the underlying structure of the network. An appealing class of models is the latent feature relational model (LFRM) (Miller et al., 2009), often also called the *overlapping* stochastic blockmodel (Latouche et al., 2011), which associates with each node a latent binary vector that can be thought of as the node's overlapping memberships to one or more latent clusters in the network.

The modeling flexibility offered by overlapping stochastic blockmodels, however, comes at a price. Inference in these models, typically performed by MCMC methods (Miller et al., 2009; Latouche et al., 2011), can be particularly challenging and is not easy to scale to networks with very large number of nodes. Moreover, many real-world networks exhibit considerably more complex interactions which may not be explained by the *flat* embeddings learned via these models. This problem can be further exacerbated due to the extreme sparsity of the observed links, and although leveraging some side information that might be available for each node can help alleviate this issue to some extent (Kim et al., 2011), this can make inference even more complex to scale to large networks (Kim et al., 2011). Besides, communities in real-world networks often tend to have inter-dependencies/hierarchical structures that are usually ignored by these single layer models.

Motivated by these limitations, we present an overarching framework that enables us to address these challenges via a unified, fully Bayesian model. Specifically, we develop a model that learns *multiple* layers of latent features of the nodes in the network, effectively learning a more expressive representation of the nodes which can better explain the interactions among the nodes in more complex networks, as compared to the existing methods. At the same time, the hierarchy of multiple layers of latent features allows imposing/exploiting the correlations among the clusters, which is usually not possible with single layer models.

Another appealing aspect of our model is its ability to in-corporate side information (given as node attributes) via a regression model that maps the node attributes to node latent features. This provides the model robustness when the network is highly sparse and/or in "cold-start" problems where a new node may not have formed links with any existing nodes and we may still want to predict its cluster memberships and/or links with the existing nodes.

[1]Yahoo! Research, New York, NY, USA  [2]CSE Department, IIT Kanpur, Kanpur, UP, India  [3]Duke University, Durham, NC, USA. Correspondence to: Changwei Hu <changweih@yahoo-inc.com>, Piyush Rai <piyush@cse.iitk.ac.in>, Lawrence Carin <lcarin@duke.edu>. This work was done when Changwei Hu was a Ph.D. student at Duke University.

Our model also enjoys excellent computational scalability. In particular, leveraging data-augmentation techniques allows us achieve full local conjugacy and enables us to develop a simple Gibbs sampler for model inference. Moreover, the inference cost in our model scales in the number of observed edges in the network, which makes it especially appealing for large real-world networks which are inherently sparse. Finally, although in this exposition, we only focus on unweighted networks (given as binary symmetric/asymmetric adjacency matrix), our framework, based on a gamma-Poisson construction, can readily be applied to weighted networks (Aicher et al., 2013) where the edges may have count-valued weights.

## 2. The Model

We denote the network being modeled as a binary adjacency matrix $\mathbf{A} \in \{0,1\}^{N \times N}$, where $N$ is the number of nodes. The network may be symmetric (undirected) or asymmetric (directed). In addition to $\mathbf{A}$, we may be also given side information associated with each node. The side information, when given, will be denoted using an $N \times D$ matrix $\mathbf{S}$, with $D$ being the number of observed features in the side information, and $s_i \in \mathbb{R}^D$ (row $i$ in $\mathbf{S}$) denoting the side information associated with node $i$.

Following the overlapping stochastic blockmodel (Latouche et al., 2011; Miller et al., 2009; Zhu, 2012) approach to statistical network modeling, we assume that each node $i$ in the network can be described as *binary* latent feature vector $z_i$ of size $K$, where $z_{ik} = 1$ if node $i$ belongs to the latent cluster/community $k$ (and 0 otherwise). Note that the model allows each node to belong to more than one cluster/community. Given the node latent features, the probability of a link $A_{ij}$ between nodes $i$ and $j$ can then be defined as a (bilinear) function of the latent features $z_i$ and $z_j$, i.e., $p(A_{ij} = 1) = g(z_i^\top \Lambda z_j)$ where $\Lambda$ is a $K \times K$ matrix, with $\Lambda_{k\ell}$ modulating the probability of link between two nodes belonging to clusters $k$ and $\ell$. Here $g$ is a function (described in Sec. 2.3)) which turns real-valued scores $z_i^\top \Lambda z_j$ into probabilities.

Unlike overlapping stochastic blockmodels (Latouche et al., 2011; Miller et al., 2009; Zhu, 2012) for relational data, however, which can only learn a single layer binary latent feature representation for the nodes in form of an $N \times K$ binary matrix $\mathbf{Z} = [z_1^\top \ldots z_N^\top]$, we present a hierarchical architecture (shown in Fig. 1) which allows learning multiple layers of latent features $\{\mathbf{Z}^{(\ell)}\}_{\ell=1}^L$ where $L$ denotes the number of layers, $\mathbf{Z}^{(\ell)} \in \{0,1\}^{N \times K_\ell}$, and $K_\ell$ is the number of latent features in layer $\ell$.

Note that our proposed framework is similar in spirit to deep sigmoid belief-nets (Neal, 1992; Gan et al., 2015), originally proposed for vector-valued observations. In contrast, our focus here is to model relational data given as pairwise observations. Moreover, our framework also allows conditioning the latent features directly on the side information using a regression model. We now describe the various components of our framework in the following subsections.

### 2.1. A Structured Hierarchical Latent Feature Model

Akin to the deep sigmoid belief-nets (Neal, 1992; Gan et al., 2015), we condition each node's latent features in layer $\ell$ on its latent features in layer $\ell + 1$ via a weight matrix $\mathbf{W}^{(\ell)} \in \mathbb{R}^{K_\ell \times K_{\ell+1}}$ (Fig. 1). Thus, for node $i$, we have

$$
\begin{aligned}
p(z_{ik}^{(\ell)} = 1) &= \pi_{ik}^{(\ell)} = \sigma((\boldsymbol{w}_k^{(\ell)})^\top z_i^{(\ell+1)} + b_k^{(\ell)}), \\
\forall k &= 1, \ldots, K_\ell, \forall \ell = 1, \ldots, L-1 \\
p(z_{ik}^{(L)} = 1) &= \pi_{ik}^{(L)} = \sigma(b_k^{(L)}) \quad \forall k = 1, \ldots, K_L
\end{aligned}
$$

where $\boldsymbol{w}_k^{(\ell)} \in \mathbb{R}^{K_{\ell+1}}$ denotes the $k$-th row of $\mathbf{W}^{(\ell)}$, and $\boldsymbol{b}^{(\ell)} = (b_1^{(\ell)}, \ldots, b_{K_\ell}^{(\ell)})$ is vector of biases. Note that $L = 1$ corresponds to a single layer model.

A key benefit of the multi-layer (2 layers or more) architecture is that the 2nd layer latent features allow modeling and leveraging the correlations among the layer 1 latent features (i.e., clusters) which directly touch the data. In contrast, a flat 1 layer model will not be able to model correlations.

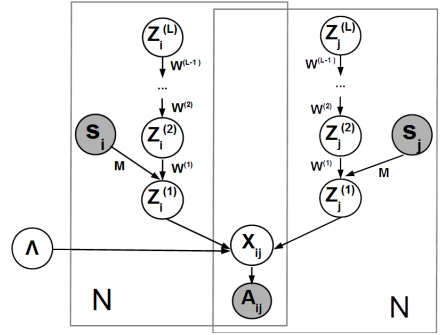

*Figure 1.* The full model with hierarchical architecture and side information. Hyperparameters not shown for brevity

### 2.2. Incorporating Side Information

If available, side information associated with the nodes in the network can be incorporated in this framework by conditioning the bottom-most layer (i.e., layer 1) latent features $\mathbf{Z}^{(1)}$ on the side information (Fig. 1). Conditional distributions of the latent features $\mathbf{Z}^{(2)}, \ldots, \mathbf{Z}^{(L)}$ in all other

Deep Generative Models for Relational Data with Side Informationlayers remain unchanged as before (Sec. 2.1), whereas the layer 1 latent features for node $i$ are now modeled as

$$p(z_{ik}^{(1)} = 1) = \sigma((\boldsymbol{w}_k^{(1)})^\top \boldsymbol{z}_i^{(2)} + \boldsymbol{m}_k^\top \boldsymbol{s}_i + b_k^{(1)}) \quad (1)$$

where $\boldsymbol{m}_k \in \mathbb{R}^D$ denotes the regression weights, which map the observed features $\boldsymbol{s}_i$ to the latent features $\boldsymbol{z}_i$. Note that although we only condition the layer 1 latent features on the side information, rest of the layers can also be conditioned on the side information in the same way.

Note that, as opposed to conditioning the link $A_{ij}$ on the side information, in our model construction we choose to condition the latent features of each node on its side information. This allows the side information to directly influence the latent features, which is useful for predicting the latent features for *new* nodes that do not have any existing links in the network. This modeling choice has also been employed in (Kim et al., 2011), an extension of mixed-membership stochastic blockmodels (Airoldi et al., 2008), where each node's cluster membership probabilities are directly conditioned on the metadata (observed features) associated with that node.

### 2.3. Generating the Network

The layer 1 latent features $\mathbf{Z}^{(1)}$ generate the observed network $\mathbf{A}$ (graphical model shown in Fig. 1). Specifically, each edge $A_{ij} \in \{0, 1\}$ is generated by thresholding a *latent* count random variable $X_{ij}$. Each of these latent counts $X_{ij}$, in turn, is defined as a summation of another set of (smaller) latent counts $X_{ijk_1k_2}$, which are defined as bilinear functions of the layer 1 latent features $\mathbf{Z}^{(1)}$. Formally,

$$A_{ij} = \mathbf{1}(X_{ij} \geq 1), X_{ij} = \sum_{k_1=1}^{K} \sum_{k_2=1}^{K} X_{ijk_1k_2}$$

$$X_{ijk_1k_2} \sim \text{Poisson}(z_{ik_1}^{(1)} \Lambda_{k_1k_2} z_{jk_2}^{(1)}) \quad (2)$$

Marginalizing out the latent counts $X_{ij}$ from Eq. 2

$$p(A_{ij} = 1) = \text{Bernoulli}\left(1 - \exp(-\boldsymbol{z}_i^{(1)\top} \Lambda \boldsymbol{z}_j^{(1)})\right)(3)$$

Note that only the bottom layer (layer 1) latent features directly touch the data layer (the observed links/non-links). The construction in Eq. 2 based on decomposing a discrete random variable into a set of latent counts has also been used previously in modeling discrete (count or binary) data (Dunson & Herring, 2005; Gopalan et al., 2014; Zhou, 2015). This construction has the appealing property that if $A_{ij} = 0$ then the associated latent counts are zero with probability one and need not be estimated during model inference. Therefore, this can lead to huge computational speed-ups for sparse data with many zeros (which is usually the case with real world networks which are very sparse). In the case of modeling relational data such as networks, this implies that the inference cost scales in the number of edges in the network, unlike other overlapping stochastic blockmodels such as LFRM (Miller et al., 2009; Zhu, 2012). These models use a logistic link function for the edges, which requires likelihood evaluations for both edges as well as non-edges. Consequently, the inference cost is quadratic in the number of nodes, making these models prohibitive for large networks. Note that the model readily applies to graphs with count-valued edges (the additional step of latent-count thresholding would not be required).

Note that a similar construction for network generation was recently employed in (Zhou, 2015). However, our framework differs from (Zhou, 2015) in a number of key ways. In particular, unlike (Zhou, 2015) in which the latent features are positive reals, in our framework the latent features are binary (in the spirit of stochastic blockmodels (Miller et al., 2009; Zhu, 2012)). The binary latent features are also crucial for a deep sigmoid belief net construction. Moreover, unlike the model in (Zhou, 2015) which cannot leverage side information, our framework allows incorporating the side information of each node in predicting the node's latent features. This capability allows our framework to work in the cold-start settings where a new node may not yet have formed any links with the existing nodes.

### 2.4. The Full Generative Model

The full generative model for the observed network $\mathbf{A}$, along with the latent variables, parameters, and hyperparameters of the model, is given below

$$A_{ij} = \mathbf{1}(X_{ij} \geq 1) \quad (4)$$

$$X_{ij} = \sum_{k_1=1}^{K} \sum_{k_2=1}^{K} X_{ijk_1k_2} \quad (5)$$

$$X_{ijk_1k_2} \sim \text{Poisson}(z_{ik_1}^{(1)} \Lambda_{k_1k_2} z_{jk_2}^{(1)}) \quad (6)$$

$$z_{ik}^{(\ell)} \sim \text{Bernoulli}(\pi_{ik}^{(\ell)}) \quad (7)$$

$$(\forall k = 1, \ldots, K_\ell, \forall \ell = 1, \ldots, L) \quad (8)$$

$$\pi_{ik}^{(\ell)} = \begin{cases} \sigma((\boldsymbol{w}_k^{(\ell)})^\top \boldsymbol{z}_i^{(\ell+1)} + \boldsymbol{m}_k^\top \boldsymbol{s}_i + b_k^{(\ell)}) \\ (\text{if } \ell = 1 \text{ and side info } \boldsymbol{s}_i \text{ available}) \\ \sigma((\boldsymbol{w}_k^{(\ell)})^\top \boldsymbol{z}_i^{(\ell+1)} + b_k^{(\ell)}) \\ (\text{if } \ell < L \text{ and side info } \boldsymbol{s}_i \text{ not available}) \\ \sigma(b_k^{(L)}) \text{ if } \ell = L \end{cases} \quad (9)$$

$$\Lambda_{k_1k_2} \sim \text{Gamma}(g_{k_1k_2}, 1/c_{k_1k_2}) \quad (10)$$

$$g_{k_1k_2} = \begin{cases} \gamma_{k_1} \gamma_{k_2} & \text{if } k_1 \neq k_2 \\ \xi \gamma_{k_1} & \text{if } k_1 = k_2 \end{cases} \quad (11)$$

$$\gamma_k \sim \text{Gamma}(\gamma_a, 1/\gamma_b), \xi \sim \text{Gamma}(\xi_a, 1/\xi_b) (12)$$

$$\boldsymbol{w}_k^{(\ell)} \sim \mathcal{N}(0, \boldsymbol{\Gamma}_{k,\ell}^{(w)}), \boldsymbol{m}_k \sim \mathcal{N}(0, \boldsymbol{\Gamma}_k^{(m)}) \quad (13)$$

To impose sparsity on the between layer connection weights $\{\boldsymbol{w}_k^{(\ell)}\}_{k,\ell=1}^{K,L}$ and the regression weights $\{\boldsymbol{m}_k\}_{k=1}^{K}$, we use automatic relevance determination (ARD) priors on these. In particular, the ARD prior on the regression weights $\boldsymbol{m}_k$ also helps in selecting the relevant features in the side information that are the most relevant in predicting the node's binary latent features.

Another appealing property of the resulting link function (Eq. 3) is that it encourages generation of networks that are inherently sparse. To see this, note that using the likelihood model given by Eq. 3 readily leads to a lower bound on the number of zeros in the $\mathbf{A}$ matrix.

**Lemma 1.** The level of sparsity of the observed network $\mathbf{A}$, as measured by the expected number of zeros in $\mathbf{A}$, i.e., $\mathbb{E}[\sum_{i,j=1}^{N} \mathbf{I}\{A_{ij} = 0\}]$ is lower bounded by $N^2 \mathbb{E}_{\mathbf{z}_i, \mathbf{z}_j, \Lambda} \left[ -\sum_{k_1, k_2=1}^{K} \Lambda_{k_1 k_2} z_{ik_1} z_{jk_2} \right] = N^2 \exp\left(-\left[\frac{\zeta \gamma_c}{\gamma_b c_{k_1 k_2}} + \frac{\gamma_a^2}{\gamma_b^2 c_{k_1 k_2}}\right] \mathbb{E}_{\mathbf{z}_i^{(1)} \mathbf{z}_j^{(1)}} \left[ z_{ik_1}^{(1)} z_{jk_2}^{(1)} \right]\right)$, where we have made use of the fact that the expectation of the term $\sum_{k_1=1}^{\infty} \sum_{k_2=1}^{\infty} \Lambda_{k_1 k_2}$ is finite. The proof of the Lemma is given in the Supplementary Material.

## 3. Inference

The model described in (4)-(13) is not conjugate. However, leveraging data augmentation techniques, we endow the model with full local conjugacy and derive a simple and efficient Gibbs sampler for model inference. The first data augmentation technique we use is based on the Poisson-multinomial equivalence (Dunson & Herring, 2005).

**Lemma 2.** Suppose that $x_1, \ldots, x_R$ are independent random variables with $x_r \sim \text{Pois}(\theta_r)$ and $x = \sum_{r=1}^{R} x_r$. Set $\theta = \sum_{r=1}^{R} \theta_r$; let $(y, y_1, \ldots, y_R)$ be another set of random variables such that $y \sim \text{Pois}(\theta)$, and $(y_1, \ldots, y_R)|y \sim \text{Mult}(y; \frac{\theta_1}{\theta}, \ldots, \frac{\theta_R}{\theta})$. Then the distribution of $\mathbf{x} = (x, x_1, \ldots, x_R)$ is the same as the distribution of $\mathbf{y} = (y, y_1, \ldots, y_R)$.

Using this equivalence, given $X_{ij}$, the smaller latent counts $X_{ijk_1k_2}$'s can be easily sampled as

$$\{X_{ijk_1k_2}\} \sim \text{Mult}\left(X_{ij}; \frac{\{z_{ik_1}^{(1)} \Lambda_{k_1 k_2} z_{jk_2}^{(1)}\}}{\sum_{k_1=1}^{K} \sum_{k_2=1}^{K} z_{ik_1}^{(1)} \Lambda_{k_1 k_2} z_{jk_2}^{(1)}}\right)$$

The second data augmentation we use is based on the Pólya-Gamma (PG) strategy (Polson et al., 2013) which allows re-expressing logistic-Bernoulli likelihoods on the $z_{ik}^{(\ell)}$'s as Gaussians and consequently allows deriving closed-form posterior updates for the between-layer weights $\mathbf{w}_k^{(\ell)}$ and the regression weights $\mathbf{m}_k$ (note that each of these are given Gaussian priors). According to the PG augmentation, given likelihoods expressible in the form $\frac{(\exp(\tau))^a}{(1+\exp(\tau))^b}$, and given Pólya-Gamma random variable random variables $\omega \sim \text{PG}(b, 0)$

$$\frac{(\exp(\tau))^a}{(1+\exp(\tau))^b} = 2^{-b} \exp(\kappa \tau) \int_0^{\infty} \exp(-\kappa \tau^2/2) p(\omega) d\omega$$

where $p(\omega)$ is the density of the PG variable, and $\kappa = a - b/2$. This result transforms the logistic-Bernoulli into a Gaussian, when conditioned on the PG random variables.

For example, for sampling $\mathbf{w}_k^{(\ell)}$, we draw $N$ Pólya-Gamma random variables $\boldsymbol{\alpha}_k^{(\ell)} = [\alpha_{1k}^{(\ell)}, \cdots, \alpha_{Nk}^{(\ell)}]$, one for each (Bernoulli-drawn) $z_{ik}^{(\ell)}$, as

$$\alpha_{ik}^{(\ell)} \sim \text{PG}(1, (\mathbf{w}_k^{(\ell)})^\top \mathbf{z}_i^{(\ell+1)} + b_k^{(\ell)})$$

where PG(.) denotes the Pólya-Gamma distribution (Polson et al., 2013). Conditioned on $\boldsymbol{\alpha}_k^{(\ell)}$, the logistic-Bernoulli likelihood on $z_{ik}^{(\ell)}$ turns into a Gaussian and consequently the posterior distribution of $\mathbf{w}_k^{(\ell)}$ will also be a Gaussian. The same data augmentation strategy is followed for sampling the regression weights $\mathbf{m}_k$, for which conditioned on the layer 1 PG variables $\boldsymbol{\alpha}_k^{(1)}$, the posterior of $\mathbf{m}_k$ is a Gaussian. The details are given in the next subsection.

### 3.1. Gibbs Sampling

Gibbs sampling for our model proceeds as follows.

**Sample $X_{ij}$**: For each nonzero observation $A_{ij} = 1$ in the network, the associated latent count $X_{ij}$ is sampled from a zero-truncated Poisson distribution as

$$X_{ij} \sim \text{Pois}_+ \left( \sum_{k_1=1}^{K_1} \sum_{k_2=1}^{K_1} z_{ik_1}^{(1)} \Lambda_{k_1 k_2} z_{jk_2}^{(1)} \right)$$

**Sample $X_{ijk_1k_2}$**: Having sampled $X_{ij}$, the latent counts, $X_{ijk_1k_2}$ can be sampled using the Poisson-multinomial equivalence (Lemma 2).

**Sample $z_{ik_1}^{(1)}$**: Layer 1 latent features $z_{ik_1}^{(1)}$ are sampled as

$$z_{ik_1}^{(1)} \sim \delta(X_{i \cdot k_1 \cdot} = 0) \text{Bern}\left(\frac{\tilde{\pi}_{ik_1}}{\tilde{\pi}_{ik_1} + 1 - \pi_{ik_1}^{(1)}}\right) + \delta(X_{i \cdot k_1 \cdot} > 0)$$

where the "marginal" latent counts are defined as summations, i.e., $X_{i \cdot k_1 \cdot} = \sum_{j>i} \sum_{k_2} X_{ijk_1k_2} + \sum_{j<i} \sum_{k_2} X_{jik_2k_1}$, $\tilde{\pi}_{ik_1} = \pi_{ik_1}^{(1)} \prod_{k_2}^{K_1} (\frac{c_{k_1 k_2}}{1+c_{k_1 k_2}})^{\Lambda_{k_1 k_2} z_{\cdot k_2}^{(1)}}$, and $z_{\cdot k_2}^{(1)} = \sum_{i=1}^{N} z_{ik_2}^{(1)}$.

**Sample $\Lambda_{k_1 k_2}$**: The latent feature interaction weights $\Lambda_{k_1 k_2}$ are sampled as

$$\Lambda_{k_1 k_2} \sim \text{Gamma}\left(X_{\cdot \cdot k_1 k_2} + g_{k_1 k_2}, \frac{1}{Q_{k_1 k_2} + c_{k_1 k_2}}\right)$$

where $Q_{k_1 k_2} = \sum_{i=1}^{I} \sum_{j=1}^{J} z_{ik_1}^{(1)} z_{jk_2}^{(1)}$, $X_{\cdot \cdot k_1 k_2} = 2^{-\delta_{k_1 k_2}} \sum_i \sum_{j>i} (X_{ijk_1k_2} + X_{ijk_2k_1})$, with $\delta_{k_1 k_2} = 1$ if $k_1 = k_2$ and $\delta_{k_1 k_2} = 0$ otherwise.

**Sample $z_{ip}^{(\ell)}$ ($\ell \geq 2$)**: We consider the update of a single $z_{ip}^{(2)}$ as an example, and assume the side information is available. $z_{ip}^{(2)}$ is updated as $z_{ip}^{(2)} \sim \text{Bern}(\sigma(d_{ip}^{(2)}))$, with $d_{ip}^{(2)} = b_p^{(2)} + (\mathbf{z}_i^{(1)})^T \mathbf{w}_p^{(1)} - \frac{1}{2} \sum_{k=1}^{K_1} \left( w_{kp}^{(1)} + \right.$

$\alpha_{ik}^{(1)}(2\psi_{ik}^{\backslash p}w_{kp}^{(1)} + (w_{kp}^{(1)})^2)\Big)$, where $\boldsymbol{w}_p^{(1)} \in \mathbb{R}^{K_1}$ is the $p$-th column in $\mathbf{W}^{(1)}$, and $\psi_{ik}^{\backslash p}$ is defined as $\psi_{ik}^{\backslash p} = \boldsymbol{m}_k^T \boldsymbol{s}_i + \boldsymbol{w}_k^T \boldsymbol{z}_i^{(2)} - z_{ip}^{(2)} w_{kp}^{(1)}$.

**Sample $\boldsymbol{w}_k^{(\ell)}$, $b_k^{(\ell)}$, $\boldsymbol{m}_k$, $\gamma_{k_1}$, $\xi$, $\boldsymbol{\Gamma}_{k,\ell}^{(\boldsymbol{w})}$ and $\boldsymbol{\Gamma}_k^{(\boldsymbol{m})}$**: For brevity, the detailed equations are provided in the Supplementary Material.

## 4. Related Work

A number of extensions have been proposed to enhance the modeling capabilities of stochastic blockmodels and its variants such as the latent feature relational model (LFRM), when applied to complex graph-structured data. In particular, in the context of LFRM, recent work on infinite latent attributes (ILA) (Palla et al., 2012) is designed to learn binary latent features for the nodes in the network, and each latent feature is further assumed to be partitioned into *disjoint* groups. ILA however cannot incorporate side information, and while ILA assumes a specific two-level representation of the nodes (via latent features in level one and clusters in level two), our model is capable of learning a more general hierarchical latent feature representation.

Stochastic blockmodels have also been extended for inferring nested communities using nested Chinese Restraurant Process (Ho et al., 2012). Such methods can learn clusters of varying granularities at multiple levels in a hierarchy. However, the focus of these class of methods is different as these methods do not learn a binary latent feature based representation unlike our model and can only learn disjoint clusterings (organized at multiple scales) of nodes.

Among methods that can incorporate side information in stochastic blockmodels, the nonparametric metadata dependent relational (NMDR) model (Kim et al., 2011) is somewhat similar in spirit to our model in the way the side information is incorporated into the model. The NMDR model builds on the nonparametric Bayesian mixed-membership stochastic blockmodel (Airoldi et al., 2008) and the side information is incorporated by conditioning the cluster membership probabilities (the weights of the sticks in the stick-breaking process) on the side information via a regression model. However, it is a single layer model, requires retrospective MCMC sampling for inference, and is difficult to scale to large networks. We use NMDR as one of the baselines in our experiments.

Among other methods that can incorporate side information in link prediction models beyond stochastic blockmodels, (Menon & Elkan, 2011) presented a number of non-probabilistic approaches based on latent space models that directly use the side information in the link prediction objective function (note that LFRM also proposes doing the same (Miller et al., 2009) to incorporate side information).

However, the embeddings are not conditioned on the side information and these models cannot predict the embedding of a new node from its side information.

Our model is also similar in spirit to the recently proposed infinite edge partition model (Zhou, 2015) (we also use it as one of our baselines in the experiments) which also uses the Bernoulli-Poisson link to model each edge. However, EPM assumes positive-valued node embeddings (given gamma priors), is limited to a single layer representation, and cannot incorporate side information.

To the best of our knowledge, none of the existing methods for network modeling can learn hierarchical latent representations of the nodes. Recently, DeepWalk (Perozzi et al., 2014) was introduced as a way to learn embeddings of nodes in a network using a skip-gram model by considering short random walks along the network and using these walks as "sentences" and nodes being "words", and learns the node embedding in a manner like learning word2vec embeddings. However, these embeddings are single layer real-valued embeddings. In addition to this, some other simultaneous development on deep learning for graph-structured, relational data include graph convolutional networks (Schlichtkrull et al., 2017) and graph variational autoencoders (Kipf & Welling, 2016).

In contrast to the aforementioned methods, our framework provides a unified model which not only learns a hierarchical, *interpretable* latent feature representation of the nodes, but also incorporates node side information via a regression model. Notably, both these enrichments are naturally formulated under a multilayer sigmoid belief-net type model architecture. Moreover, the model is simple to do inference on, and can easily scale to massive, sparse networks (with binary as well as count-valued edges).

## 5. Experiments

We consider three instances of our hierarchical latent feature model (**HLFM**): one-layer HLFM, two-layer HLFM, and two-layer HLFM with side information. While our framework straightforwardly extends to more than two layers, we specifically focus our experimental analysis to consider the single and two layer cases (with/without side information), to carefully explicate the advantage of our model in: (1) going from flat to hierarchical latent features, and (2) the advantage of incorporate the side information, especially when the network is highly sparse.

We apply our model on several benchmark relational data sets, and compare with three state-of-the-art methods for stochastic blockmodeling and link prediction as baseline, including stochastic blockmodels based methods that can also incorporate side information. Our baselines include:

- Hierarchical Gamma Process Edge Partition Model

  (**HGP-EPM**) (Zhou, 2015): This is a state-of-the-art, highly scalable Bayesian model for learning overlapping communities. The model is based on learning non-negative embeddings for each node.
- Community-Affiliation Graph Model (**AGM**) (Yang & Leskovec, 2012): This model is an overlapping community detection model based on learning a binary latent feature vector (akin to our approach and latent feature relational models (Miller et al., 2009)).
- Nonparametric Metadata Dependent Relational Model (**NMDR**) (Kim et al., 2011): This model is based on the nonparametric Bayesian mixed-membership blockmodel and, in the same spirit as our model, allows conditioning a node's cluster memberships on metadata associated with that node.

### 5.1. Data Sets

We consider seven real-world data sets, with five data sets associated with side information, and the remaining two having no side information. The description of each data set (and the associated side information) is given below:

**Protein230:** This data set contains information about protein-protein interactions of 230 proteins, with 595 edges. This network has no side information.

**NIPS234:** Coauthor network consists of the top 234 authors in NIPS 1-17 conferences in terms of the number of publications, as studied in (Miller et al., 2009). There are 598 edges. This network has no side information.

**Conflicts:** Network of military disputes between countries in year 1990-2000 (Ghosn et al., 2004). The graph is symmetric, i.e., two countries have a link if either initiated conflict with the other. There are 130 countries and 320 edges. Each country has 3 features: GDP, population, and polity.

**Facebook:** User-user interactions extracted from Facebook social network (McAuley & Leskovec, 2012). There are 228 users from 14 ego-network communities. Each user is associated with 92 profile information features (e.g., age, gender, education).

**Metabolic:** Metabolic pathway interaction data for Saccharomyces cerevisiae provided in the KEGG/PATHWAY database (Yamanishi et al., 2005). There are 668 nodes in total. Each node is associated with three sets of features: phylogenetic information (157 features), gene expression information (145 features), and gene location (23 features).

**NIPS_1-17:** NIPS coauthorship network containing 2865 authors, and 9466 edges. For this dataset, we also know what words each author used in their publications. We decompose the author-word matrix using SVD, and introduce first 100 SVD-based author features as side information.

**CiteSeer**: A citation network consisting of 3312 scientific publications from six categories: agents, AI, databases, human computer interaction, machine learning, and information retrieval. The side information for the dataset is the category label for each paper which is converted into a one-hot representation.

We evaluate our model on both quantitative tasks (in its ability to predict missing links in the network) as well as qualitative tasks (interpreting the inferred clusters).

### 5.2. Predicting Held-out Links

We use Area Under the ROC Curve (AUC) to evaluate our model and the other baselines on the task of link prediction. For the two data sets without side information (Protein230 and NIPS234), we hold out $20\%$ data as our test data. For the remaining five data sets, we hold out $80\%$ data as our test data as we were interested in highly missing data regimes to investigate how much the side information is benefitting in such difficult cases.

The shrinkage priors used in our model and the other baselines can automatically prune out the unnecessary latent features. We set $K$ to a large enough number ($K = 100$) so that all models are evaluated with sufficient number of latent features. Our models and the other baselines (except HGP-EPM) are run with 1000 burn-in iterations, and another 1000 iterations for sample collection. For the HGP-EPM baseline, we use the default setting from (Zhou, 2015) and run their model for 3000 burn-in and 1000 collection iterations. The samplers are initialized randomly. Each experiment is repeated 5 times with different training and test splits and averaged results are reported.

Table 1 reports the results on the two data sets that do not have side information and Table 2 reports the results on the other four data sets with side information. On the data sets with side information, Figure 2 separately compares the three variants of our model: the model with one layer, two layers, and two layers with side information.

As shown in Table 1, our two layer model outperforms all the other methods. Also note that, on NIPS234, the one layer model is outperformed by HGP-EPM and performs comparably to AGM (which like our model learns binary latent feature for each model). However, there is a marked improvement in the performance when using the two layer model and the model ourperforms all the baselines by a significant margin. This shows the benefit of the better and more expressive latent features learned by the hierarchy in our model, even when no side information is available.

Table 2 shows the results in the presence of side information. Except for Conflicts data, where our model gets outperformed by HGP-EPN, our two layer model with side information significantly outperforms the baselines on most of the data sets. In particular, our model yields better AUC scores than the other best performing baseline

NMDR (Kim et al., 2011) which can, like our model, incorporate side information. The better performance of our model can be attributed to a combination of several factors, e.g., (1) unlike NMDR, our model allows overlapping membership to multiple clusters (and multiple layers of latent features) leading to more expressive latent features; and (2) inference is simpler in our model, which leads to better mixing of the sampler.

It is interesting to note that neither NMDR nor the two-layer HLFM with side information outperform HGP-EPM on Conflicts data in terms of the link prediction performance. This is probably because the side-information is too simple and not very informative for link prediction.

In Figure 2, we also separately compare the three variants of our model: the model with one layer, two layers, and two layers with side information. As the figure shows, the two layer model usually performs better than the one layer model, and incorporating the side information leads to further improvements in the AUC scores, with the strength of improvement depending on how informative the side information is in predicting the latent features of the nodes.

Table 1. AUC scores on Protein230 and NIPS234

|  | HGP-EPM | AGM | NMDR | HLFM $\ell = 1$ | HLFM $\ell = 2$ |
|---|---|---|---|---|---|
| Protein230 | 0.942 | 0.868 | 0.826 | 0.923 | **0.956** |
| NIPS234 | 0.939 | 0.842 | 0.796 | 0.823 | **0.951** |

Table 2. AUC scores on data sets with side information (Note: NMDR was infeasible to run on the NIPS_1-17 and CiteSeer data in a reasonable amount of time)

|  | HGP-EPM | AGM | NMDR | HLFM $\ell = 2$ side-info |
|---|---|---|---|---|
| Conflicts | **0.890** | 0.722 | 0.810 | 0.856 |
| Facebook | 0.868 | 0.726 | 0.890 | **0.896** |
| Metabolic | 0.744 | 0.672 | 0.763 | **0.828** |
| NIPS_1-17 | 0.720 | 0.566 | NA | **0.772** |
| CiteSeer | 0.868 | 0.776 | NA | **0.919** |

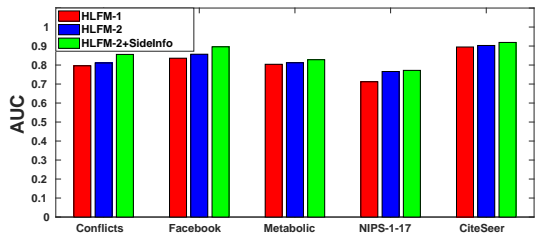

Figure 2. Comparing the three variants of our model on the data sets with side information.

### 5.3. Qualitative Analyses via Inferred Latent Features

The node embedding learned by our model (with or without side information) can be used for qualitative analyses. Note that each column of the binary latent feature matrix $\mathbf{Z}^{(\ell)}$ represents a cluster of nodes in the network. Essentially, the nonzero entries in each column of the matrix correspond to the nodes that belong to a cluster in layer $\ell$.

We use $\mathbf{Z}^{(\ell)}$ to present clustering results for the NIPS234 and Conflicts datasets in Table 3 and 4. In Table 3, showing results on the NIPS234 data, note that some authors (e.g., Michael Jordan) are inferred as belonging to more than one cluster (since the model allows overlapping clusters).

Table 3. NIPS234 - Clusters of representative authors in layer 1

| Cluster | Author |
|---|---|
| Probabilistic Modeling | Sejnowski T, Jordan M, Hinton G, Williams C, Smyth P, Frey B J, Ghahramani Z, Zemel R |
| Kernels & Learning Theory | Jordan M, Scholkopf B, Vapnik V, Shawe-Taylor J, Smola A, Platt J, Bousquet O, Smola A J |
| Cognitive Neuroscience | Touretzky D, Koch C, Mozer M, Baldi P, Moore A, Bower J, Mead C, DeWeerth S, Personnaz L |

Likewise, Table 4, shows the results on Conflicts data, with the inferred clusters of countries. To further show the discovered clusters at multiple layers and the inter-relationships between clusters: (1) In Table 4, we show the learned clusters of countries in layer 1 and layer 2; (2) In Figure 4, we show the inferred correlation-based pairwise similarities between the layer 1 clusters. To compute these correlations, we use the between-layer weights $\boldsymbol{w}_k^{(\ell)}$ as the feature vector for the $k$-th cluster (of layer $\ell = 1$) and use cosine similarity between the feature vectors of each pair of clusters.

Table 4. Conflicts Data - Country clusters in layer 1 & 2

| Cluster | Country |
|---|---|
| 3 (layer 1) | Angola, South Africa, Swaziland, Zambia |
| 4 (layer 1) | Dem. Rep. Congo, Lesotho, Mozambique |
| 1 (layer 1) | Egypt, Ghana, Guinea, Iraq, Jordan, Liberia, Libya, Niger, Nigeria, Syria |
| 10 (layer 1) | Cameroon, Ivory Coast, Chad, Iraq, Israel, Jordan, Liberia, Namibia, Sierra Leone, Sudan |
| 3 (layer 2) | Hungary, Italy, Netherlands, Iraq, Sudan, Yemen, North Korea, Malaysia |

Table 5. Computational time (seconds/iteration) comparision (Note: Two-layer HLFM with side inforamtion was infeasible to run on NIPS234 and Protein230 for lack of side information.)

|  | HGP-EPM | AGM | NMDR | HLFM $\ell = 1$ | HLFM $\ell = 2$ | HLFM $\ell = 2$ side-info |
|---|---|---|---|---|---|---|
| NIPS234 | 0.191 | 0.003 | 3.08 | 0.255 | 0.303 | NA |
| Protein230 | 0.210 | 0.003 | 3.24 | 0.299 | 0.380 | NA |
| Conflicts | 0.023 | 0.019 | 0.31 | 0.023 | 0.026 | 0.030 |
| Metabolic | 0.192 | 0.188 | 6.70 | 0.237 | 0.255 | 0.306 |

From the left plot of Figure 4, it can be observed, for example, that layer 1 clusters 3 and 4 have a high similarity, and clusters 1 and 10 have a high similarity. Looking at these four clusters, which are also shown in Table 4, we find that the countries in each of these layer 1 clusters are usually bordering countries (as shown in the right plot of Figure 4) having military disputes or other types of bilateral relations (e.g., military aid). Interestingly, unlike layer 1 clusters, the countries grouped together in layer 2 clusters are not necessarily related by the virtue of being geographically close. As seen in Table 4, the clusters in layer 2 (e.g., cluster 3) are more coarse-grained, and can be regarded as a "super group" of clusters . Such clusters consist of countries from multiple geographic regions, such as Europe, Middle East and Asia, some of which are known to be related via some military disputes, despite not being geographically close. For example, during the Gulf war (1991, recorded in Conflicts data between 1990-2000), Iraq (Middle East) was involved disputes with the coalition members which included countries like Hungary, Italy, Netherlands (Europe). Interestingly all these countries are grouped together in cluster 3 of layer 2. This analysis demonstrates that the multilayer architecture of our model not only yields significantly improved link-prediction accuracies but also enables us in gaining better insights into the data by means of more interpretable latent features and clusterings, which may be useful in their own right in many applications.

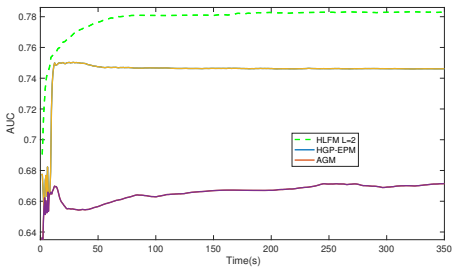

Figure 3. Comparison of the AUC convergence of HLFM ($\ell$=2), HGP-EPM, and AGM on Metabolic data.

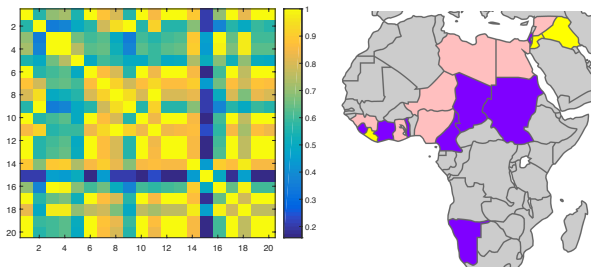

Figure 4. (Best seen in color) Left: inferred pairwise similarities between layer 1 clusters; Right: clusters 1 and 10 in layer 1. Countries in pink and purple colors are assigned to cluster 1 and 10 respectively, and countries in yellow assigned to both clusters.

### 5.4. Computational Efficiency And Convergence

We also perform an experiment to assess the computational efficiency of our framework. We compare (on four data sets) the run time of the three variants of our model (one layer, two layers, and two layers with side information) with the run times of the NMDR (Kim et al., 2011), AGM(Yang & Leskovec, 2012) and HGP-EPM (Zhou, 2015), all of which are state-of-the-art community detection/link prediction methods. All the models are implemented in MATLAB and were run on a standard machine with 2.40GHz processor and 16GB RAM. Our inference routines are based on batch Gibbs sampling. The per-iteration computation times are shown in Table 5.

As shown in Table 5, our models have very small per-iteration run times which are comparable with the other baselines. Among all the methods compared, note that AGM has smallest computational cost. This is due to the simplicity of the model (however it also gives the lowest AUC scores on all the data sets). Besides AGM, our models have run times that are comparable to the baseline HGP-EPM (which is a single layer model and cannot incorporate side information) and are considerably faster than the other baseline NMDR which, although capable of incorporating side information, is computationally much more expensive as compared with our models.

We also compare (Figure 3) the empirical convergence of the various models on the Metabolic data (80/20 training/testing split). As the figure shows, the convergence time for our two-layer HLFM is comparable with HGP-EPM model, while AGM takes the longest to converge.

## 6. Conclusion

We presented a deep generative model for relational data for which side information may also be available for each node. Our model enriches the latent feature relational models for networks using a hierarchical structure, and allows incorporating side information seamlessly via a regression model. To the best of our knowledge, ours is the first framework that extends overlapping stochastic blockmodels to a deep architecture. A key benefit of the deep architecture (even with 2 hidden layers) is that the layer 2 latent features allow modeling/leveraging correlations among layer 1 latent features/clusters which directly touch the data. A flat model will not be able to leverage such correlations. The modeling flexibility is also accompanied by simplicity of inference. In particular, leveraging data augmentation schemes, the model enjoys full local conjugacy and admits efficient inference via a simple Gibbs sampler. Networks/graphs with binary as well as count-weighted edges can be analyzed using our model (by replacing the truncated Poisson likelihood by a Poisson likelihood). The model can be easily scaled up even further using online Bayesian inference, and by leveraging recognition models (Kingma & Welling, 2013) for fast inference of the latent features. Another possible extension of the model will be in modeling multi-relational data, such as knowledge-graphs (Schlichtkrull et al., 2017; Hu et al., 2016).

**Acknowledgements:** This research was supported in part by ARO, DARPA, DOE, NGA, ONR and NSF. Piyush Rai also acknowledges support from IBM Faculty Award, DST-SERB Early Career Research Award, Dr. Deep Singh and Daljeet Kaur Faculty Fellowship, and the Research-I Foundation, IIT Kanpur.

# References

Aicher, Christopher, Jacobs, Abigail Z, and Clauset, Aaron. Adapting the stochastic block model to edge-weighted networks. *arXiv preprint arXiv:1305.5782*, 2013.

Airoldi, Edoardo M, Blei, David M, Fienberg, Stephen E, and Xing, Eric P. Mixed membership stochastic blockmodels. *JMLR*, 2008.

Dunson, David B and Herring, Amy H. Bayesian latent variable models for mixed discrete outcomes. *Biostatistics*, 6(1):11–25, 2005.

Fortunato, Santo. Community detection in graphs. *Physics reports*, 486(3):75–174, 2010.

Gan, Zhe, Henao, Ricardo, Carlson, David E, and Carin, Lawrence. Learning deep sigmoid belief networks with data augmentation. In *AISTATS*, 2015.

Ghosn, Faten, Palmer, Glenn, and Bremer, Stuart. The mid3 data set, 1993-2001: Procedures, coding rules, and description. *Conflict Management and Peace Science*, 2004.

Goldenberg, Anna, Zheng, Alice X, Fienberg, Stephen E, and Airoldi, Edoardo M. A survey of statistical network models. *Foundations and Trends® in Machine Learning*.

Gopalan, Prem, Ruiz, Francisco J, Ranganath, Rajesh, and Blei, David M. Bayesian nonparametric poisson factorization for recommendation systems. In *AISTATS*, pp. 275–283, 2014.

Ho, Qirong, Parikh, Ankur P, and Xing, Eric P. A multiscale community blockmodel for network exploration. *Journal of the American Statistical Association*, 107(499):916–934, 2012.

Hoff, Peter D, Raftery, Adrian E, and Handcock, Mark S. Latent space approaches to social network analysis. *JASA*, 2002.

Hu, Changwei, Rai12, Piyush, and Carin, Lawrence. Topic-based embeddings for learning from large knowledge graphs. In *Proceedings of the 19th International Conference on Artificial Intelligence and Statistics*, pp. 1133–1141, 2016.

Kim, Dae Il, Hughes, Michael, and Sudderth, Erik. The nonparametric metadata dependent relational model. In *ICML*, 2011.

Kingma, Diederik P and Welling, Max. Auto-encoding variational bayes. *arXiv preprint arXiv:1312.6114*, 2013.

Kipf, Thomas N and Welling, Max. Variational graph auto-encoders. *arXiv preprint arXiv:1611.07308*, 2016.

Latouche, Pierre, Birmelé, Etienne, and Ambroise, Christophe. Overlapping stochastic block models with application to the french political blogosphere. *The Annals of Applied Statistics*, 2011.

McAuley, Julian and Leskovec, Jure. Learning to discover social circles in ego networks. In *NIPS*, 2012.

Menon, Aditya Krishna and Elkan, Charles. Link prediction via matrix factorization. In *Machine Learning and Knowledge Discovery in Databases*. 2011.

Miller, Kurt, Griffiths, Thomas, and Jordan, Michael. Nonparametric latent feature models for link prediction. *NIPS*, 2009.

Neal, Radford M. Connectionist learning of belief networks. *Artificial intelligence*, 56(1):71–113, 1992.

Nowicki, Krzysztof and Snijders, Tom A B. Estimation and prediction for stochastic blockstructures. *JASA*, 2001.

Palla, Konstantina, Knowles, David, and Ghahramani, Zoubin. An infinite latent attribute model for network data. In *ICML*, 2012.

Perozzi, Bryan, Al-Rfou, Rami, and Skiena, Steven. Deepwalk: Online learning of social representations. In *KDD*, 2014.

Polson, Nicholas G, Scott, James, and Windle, Jesse. Bayesian inference for logistic models using pólya–gamma latent variables. *Journal of the American Statistical Association*, 108(504):1339–1349, 2013.

Schlichtkrull, Michael, Kipf, Thomas N, Bloem, Peter, Berg, Rianne van den, Titov, Ivan, and Welling, Max. Modeling relational data with graph convolutional networks. *arXiv preprint arXiv:1703.06103*, 2017.

Schmidt, Mikkel N and Morup, Morten. Nonparametric bayesian modeling of complex networks: An introduction. *Signal Processing Magazine, IEEE*, 30(3), 2013.

Yamanishi, Yoshihiro, Vert, Jean-Philippe, and Kanehisa, Minoru. Supervised enzyme network inference from the integration of genomic data and chemical information. *Bioinformatics*, 2005.

Yang, Jaewon and Leskovec, Jure. Community-affiliation graph model for overlapping network community detection. In *ICDM*, 2012.

Zhou, Mingyuan. Infinite edge partition models for overlapping community detection and link prediction. In *AISTATS*, 2015.

Zhu, Jun. Max-margin nonparametric latent feature models for link prediction. In *ICML*, 2012.

# Supplementary Material: Deep Generative Models for Relational Data with Side Information

## 1. Proof of Lemma 1

We can compute $\mathbb{E}[\mathbf{I}\{A_{ij}=0\}]$ as

$$\mathbb{E}[\mathbf{I}\{A_{ij}=0\}] = p(X_{ij}=0)$$

$$= \mathbb{E}_{\boldsymbol{z}_i,\boldsymbol{z}_j,\Lambda}\left[\prod_{k_1,k_2}^{K} p(X_{ij}=0|z_{ik_1},z_{jk_2},\Lambda_{k_1k_2})\right]$$

$$= \mathbb{E}_{\boldsymbol{z}_i,\boldsymbol{z}_j,\Lambda}\left[\prod_{k_1,k_2=1}^{K} \exp(-\Lambda_{k_1k_2}z_{ik_1}z_{jk_2})\right]$$

$$\geq \exp\left(\mathbb{E}_{\boldsymbol{z}_i,\boldsymbol{z}_j,\Lambda}\left[\log\prod_{k_1,k_2=1}^{K}\exp(-\Lambda_{k_1k_2}z_{ik_1}z_{jk_2})\right]\right)$$

$$= \mathbb{E}_{\boldsymbol{z}_i,\boldsymbol{z}_j,\Lambda}\left[-\sum_{k_1,k_2=1}^{K}\Lambda_{k_1k_2}z_{ik_1}z_{jk_2}\right] \quad (1)$$

where the inequality step follows from Jensen's inequality. Following Lemma 1 in (Zhou, 2015), we have $\mathbb{E}\left[\sum_{k_1,k_2=1}^{K}\Lambda_{k_1k_2}\right] = \frac{\zeta\gamma_c}{\gamma_b c_{k_1k_2}} + \frac{\gamma_a^2}{\gamma_b^2 c_{k_1k_2}}$. Then the last line in Equation (1) can be written as

$$\mathbb{E}_{\boldsymbol{z}_i,\boldsymbol{z}_j,\Lambda}\left[-\sum_{k_1,k_2=1}^{K}\Lambda_{k_1k_2}z_{ik_1}z_{jk_2}\right] \quad (2)$$

$$= \exp\left(-\left[\frac{\zeta\gamma_c}{\gamma_b c_{k_1k_2}} + \frac{\gamma_a^2}{\gamma_b^2 c_{k_1k_2}}\right]\mathbb{E}_{\boldsymbol{z}_i^{(1)}\boldsymbol{z}_j^{(1)}}\left[z_{ik_1}^{(1)}z_{jk_2}^{(1)}\right]\right)$$

Based on Equation (1) and (3), the expected number of zeros in $\mathbf{A}$ is lower bounded by

$$\mathbb{E}[\sum_{i,j=1}^{N}\mathbf{I}\{A_{ij}=0\}] \geq N^2\mathbb{E}_{\boldsymbol{z}_i,\boldsymbol{z}_j,\Lambda}\left[-\sum_{k_1,k_2=1}^{K}\Lambda_{k_1k_2}z_{ik_1}z_{jk_2}\right]$$

$$= N^2\exp\left(-\left[\frac{\zeta\gamma_c}{\gamma_b c_{k_1k_2}} + \frac{\gamma_a^2}{\gamma_b^2 c_{k_1k_2}}\right]\mathbb{E}_{\boldsymbol{z}_i^{(1)}\boldsymbol{z}_j^{(1)}}\left[z_{ik_1}^{(1)}z_{jk_2}^{(1)}\right]\right) \quad (3)$$

## 2. HYPERPARAMETER INFERENCE

We sample $\boldsymbol{w}_k^{(\ell)}$, $b_k^{(\ell)}$ and $\boldsymbol{m}_k$ leveraging the Pólya-Gamma augmentation (Polson et al., 2013). This enables us to derive the Gibbs sampler updates for the hyper-parameters $\gamma_{k_1}$, $\xi$, $\boldsymbol{\Gamma}_{k,\ell}^{(\boldsymbol{w})}$ and $\boldsymbol{\Gamma}_k^{(\boldsymbol{m})}$, in closed form.

**Sample $\boldsymbol{w}_k^{(\ell)}$ and $b_k^{(\ell)}$:** We consider the update of layer-1 weights $\boldsymbol{w}_k^{(1)}$ as an example, and assume the side information is available (which is the more general case). Weights for the other layers can be sampled in a similar manner.

Given the Pólya-Gamma auxiliary variables $\boldsymbol{\alpha}_k^{(1)}$, the posterior for $\boldsymbol{w}_k^{(1)}$ will be $\boldsymbol{w}_k^{(1)} \sim \mathcal{N}(\boldsymbol{\mu}_k^{(\boldsymbol{w})}, \mathbf{V}_k^{(\boldsymbol{w})})$, where

$$\boldsymbol{\mu}_k^{(\boldsymbol{w})} = \mathbf{V}_k^{(\boldsymbol{w})}(\mathbf{Z}^{(2)})^T(\boldsymbol{z}_k^{(2)} - \tfrac{1}{2}\mathbf{1}_N - \mathrm{diag}(\boldsymbol{\alpha}_k^{(1)})(\mathbf{S}\boldsymbol{m}_k + b_k^{(1)}\mathbf{1}_N))$$

$$\mathbf{V}_k^{(\boldsymbol{w})} = ((\mathbf{Z}^{(2)})^T\mathrm{diag}(\boldsymbol{\alpha}_k^{(1)})\mathbf{Z}^{(2)} + (\boldsymbol{\Gamma}_{k,\ell}^{(\boldsymbol{w})})^{-1})^{-1}$$

In the above, $\mathbf{1}_N$ is a vector of length $N$ with all entries being 1, and $\boldsymbol{\alpha}_k^{(1)} \in \mathbb{R}_+^N$, each entry $\alpha_{ik}^{(1)}$ is drawn from the Pólya-Gamma distribution

$$\alpha_{ik}^{(1)} \sim \mathrm{PG}(1, \boldsymbol{m}_k^T\boldsymbol{s}_i + (\boldsymbol{w}_k^{(1)})^\top \boldsymbol{z}_i^{(2)} + b_k^{(1)})$$

Conditioned on these PG variables, the posterior over $b_k^{(\ell)}$ will also be a Gaussian.

**Sample $\boldsymbol{m}_k$:** Akin to the way we sample $\boldsymbol{w}_k^{(\ell)}$, the side information based regression weights $\boldsymbol{m}_k$ can also be sampled using the Pólya-Gamma scheme (using the layer 1 PG variables $\boldsymbol{\alpha}_k^{(1)}$). The posterior will be a Gaussian $\boldsymbol{m}_k \sim \mathcal{N}(\boldsymbol{\mu}_k^{(\boldsymbol{m})}, \mathbf{V}_k^{(\boldsymbol{m})})$, where

$$\boldsymbol{\mu}_k^{(\boldsymbol{m})} = \mathbf{V}_k^{(\boldsymbol{m})}\mathbf{S}^T(\boldsymbol{z}_k^{(2)} - \tfrac{1}{2}\mathbf{1}_N - \mathrm{diag}(\boldsymbol{\alpha}_k^{(1)})(\mathbf{Z}^{(2)}\boldsymbol{w}_k^{(1)} + b_k^{(1)}\mathbf{1}_N))$$

$$\mathbf{V}_k^{(\boldsymbol{m})} = ((\mathbf{S}^T\mathrm{diag}(\boldsymbol{\alpha}_k^{(1)})\mathbf{S} + (\boldsymbol{\Gamma}_k^{(\boldsymbol{m})})^{-1})^{-1}$$

**Sample $\gamma_{k_1}$:** $\gamma_{k_1}$ can be sampled as

$$\gamma_{k_1} \sim \mathrm{Gamma}(\gamma_a + \ell_{k_1k_2}, \frac{1}{\gamma_b - \sum_{k_2}\xi^{\delta_{k_1k_2}}\gamma_{k_2}^{1-\delta_{k_1k_2}}\ln(\frac{c_{k_1k_2}}{Q_{k_1k_2}+c_{k_1k_2}})})$$

where $\ell_{k_1\cdot} = \sum_{k_2}\ell_{k_1k_2}$ with $\ell_{k_1k_2}$ drawn from the Chinese Restaurant Table (CRT) distribution (Zhou, 2015)

$$\ell_{k_1k_2} \sim \mathrm{CRT}(X_{\cdot\cdot k_1k_2}, g_{k_1k_2})$$

**Sample $\xi$:** The hyperparameter $\xi$ can be sampled as

$$\xi \sim \mathrm{Gamma}(\xi_a + \sum_k \ell_{kk}, \frac{1}{\xi_b - \sum_k \gamma_k \ln(\frac{c_{kk}}{Q_{kk}+c_{kk}})})$$

. Correspondence to: Changwei Hu <changweih@yahoo-inc.com>, Piyush Rai <piyush@cse.iitk.ac.in>, Lawrence Carin <lcarin@duke.edu>.

Supplementary Material: Deep Generative Models for Relational Data with Side Information

**Sample $\Gamma^{(w)}_{k,\ell}, \Gamma^{(m)}_k$**: Each diagonal entry of the precision matrix $\Gamma^{(w)}_{k,\ell}$ is sampled as

$$\Gamma^{(w)}_{k,\ell} \sim \text{Gamma}(a + \frac{K_{\ell+1}}{2}, \frac{1}{\text{diag}((b + 0.5(w^{(\ell)}_k)^T w^{(\ell)}_k)\mathbf{1}_{K_{\ell+1}})})$$

where $a$ and $b$ are the scale and rate parameters for the prior of $\Gamma^{(w)}_{k,\ell}$ respectively. $\Gamma^{(m)}_k$ can be sampled similarly.

# References

Polson, Nicholas G, Scott, James, and Windle, Jesse. Bayesian inference for logistic models using pólya–gamma latent variables. *Journal of the American Statistical Association*, 108(504):1339–1349, 2013.

Zhou, Mingyuan. Infinite edge partition models for overlapping community detection and link prediction. In *AISTATS*, 2015.